\pdfoutput=1

\documentclass[11pt]{article}

\usepackage{acl}

\usepackage{times}
\usepackage{latexsym}

\usepackage[T1]{fontenc}

\usepackage{microtype}

\usepackage{multirow}
\usepackage{graphicx}
\usepackage{booktabs}
\usepackage{amsmath,amssymb}
\usepackage{color}
\usepackage{booktabs}

\title{Dial2vec: Self-Guided Contrastive Learning \\of Unsupervised Dialogue Embeddings}

\author{
    Che Liu,
    Rui Wang,
    Junfeng Jiang,
    Yongbin Li\thanks{\quad Corresponding author.},
    Fei Huang \\
    DAMO Academy, Alibaba Group \\
    \texttt{ \small 
    \{liuche.lc,wr224079,jiangjunfeng.jjf,shuide.lyb,f.huang\}@alibaba-inc.com
    }
}

\begin{document}
\maketitle
\begin{abstract}
In this paper, we introduce the task of learning unsupervised dialogue embeddings.
Trivial approaches such as combining pre-trained word or sentence embeddings and encoding through pre-trained language models (PLMs) have been shown to be feasible for this task.
However, these approaches typically ignore the conversational interactions between interlocutors, resulting in poor performance.
To address this issue, we proposed a self-guided contrastive learning approach named dial2vec.
Dial2vec considers a dialogue as an information exchange process.
It captures the conversational interaction patterns between interlocutors and leverages them to guide the learning of the embeddings corresponding to each interlocutor.
The dialogue embedding is obtained by an aggregation of the embeddings from all interlocutors.
To verify our approach, we establish a comprehensive benchmark consisting of six widely-used dialogue datasets.
We consider three evaluation tasks: domain categorization, semantic relatedness, and dialogue retrieval.
Dial2vec achieves on average 8.7, 9.0, and 13.8 points absolute improvements in terms of purity, Spearman's correlation, and mean average precision (MAP) over the strongest baseline on the three tasks respectively.
Further analysis shows that dial2vec obtains informative and discriminative embeddings for both interlocutors under the guidance of the conversational interactions and achieves the best performance when aggregating them through the interlocutor-level pooling strategy.
All codes and data are publicly available at \href{https://github.com/AlibabaResearch/DAMO-ConvAI/tree/main/dial2vec}{https://github.com/AlibabaResearch/DAMO-ConvAI/tree/main/dial2vec}.
\end{abstract}

\section{Introduction}
Dialogue embedding, as a critical prerequisite of semantically understanding a dialogue, has been a central issue in dialogue-related research such as dialogue clustering \cite{shi2018auto,lv2021task}, conversational sentiment analysis \cite{wang2020sentiment,lv2021task}, context-dependent text-to-SQL \cite{hui2021dynamic,wang2022proton}, and dialogue summarization \cite{liu2019topic,liu2021controllable}.
Trivial unsupervised approaches generally encode dialogues by combining their pre-trained word or sentence embeddings \cite{pennington2014glove, reimers2019sentence} or using PLMs \cite{wu2020tod, bao2020plato, he2022space, he2022unified, he2022galaxy}.
However, such methods are not specifically designed for dialogues and thus fail to adequately capture the key conversational information.
In this paper, we formally introduce the task of learning unsupervised dialogue embeddings, which aims to learn dialogue embeddings that can well reflect conversational semantics without any additional manual annotations.

Previous studies have extensively demonstrated the importance of encoding token-level interactions for learning semantic textual embeddings.
However, for dialogue embedding, encoding interlocutor-level interactions is also essential but is overlooked in trivial approaches.
\begin{figure}[hbt]
\setlength{\belowcaptionskip}{-.2cm}
  \centering
  \includegraphics[width=0.9\linewidth]{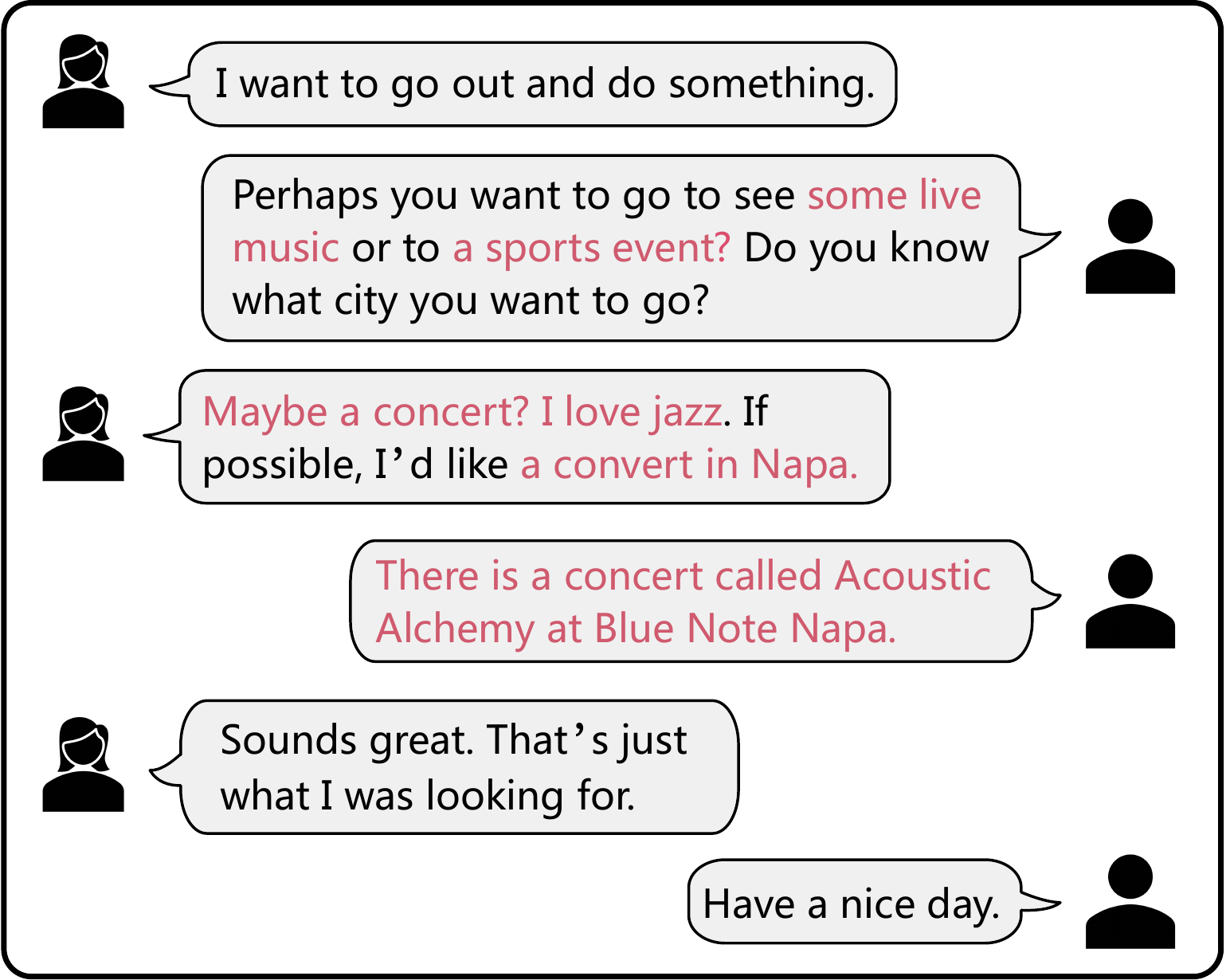}
  \caption{A dialogue from the SGD dataset.}
  \label{figure:example}
\end{figure}
Figure \ref{figure:example} shows an example.
We highlight the significant interaction patterns between the interlocutors with red color.
As we can see, although these patterns only appear in three utterances, they highly represent the key conversational semantics (e.g., topics) and are more important than the other parts (e.g., greetings and chit-chats).
We hold that capturing and leveraging them is one of the keys to learning high-quality unsupervised dialogue embeddings.

In this work, we propose dial2vec, a self-guided contrastive learning approach to solve the proposed task.
Dial2vec considers a dialogue as an information exchange process between the two interlocutors and learns embeddings for both interlocutors with the help of each other.
Specifically, dial2vec firstly encodes a dialogue through a PLM and assigns each interlocutor a self-representation by masking the non-corresponding positions in the encoding outputs.
Then it calculates a matching matrix via the token-level dot-product operation between the two self-representations, obtaining a cross-representation for each interlocutor.
Finally, the two cross-representations are leveraged as guidance to help the two self-representations gradually learn the interlocutor-level interaction-aware information and eliminate the interaction-free information during the training procedure.

To verify our model, we build a comprehensive benchmark comprising a total of 98,879 dialogues by introducing six widely-used dialogue datasets, including BiTOD \cite{lin2021bitod}, Doc2dial \cite{feng2020doc2dial}, MetalWOZ \cite{lee2019multi-domain}, MultiWOZ \cite{eric2019multiwoz}, Self-dialogue \cite{fainberg2018talking}, and SGD \cite{rastogi2020towards}.
Each dataset consists of thousands of dialogues, where each dialogue is provided with a domain label (e.g., hotel booking and movie).
We leverage these labels and design three evaluation tasks: domain categorization, semantic relatedness, and dialogue retrieval.
We categorize them into intrinsic and extrinsic tasks according to their different focus.

Experimental results on this benchmark show that dial2vec outperforms the baselines by a substantial margin.
Compared with the strongest baseline, dial2vec achieves on average 8.7, 9.0, and 13.8 points absolute improvements in terms of purity, Spearman's correlation, and mean average precision (MAP) on the three tasks respectively.
We also conduct experiments with the single interlocutor's embeddings, their aggregation strategies, and the overall dialogue embedding distributions to study how dial2vec achieves such advanced performance.
The results demonstrate that dial2vec learns both informative and discriminative embeddings for the two interlocutors and achieves the best performance when combining them through the proposed interlocutor-level pooling aggregation strategy.

\begin{figure*}[htb]
\setlength{\belowcaptionskip}{-.4cm}
  \centering
  \includegraphics[width=\linewidth]{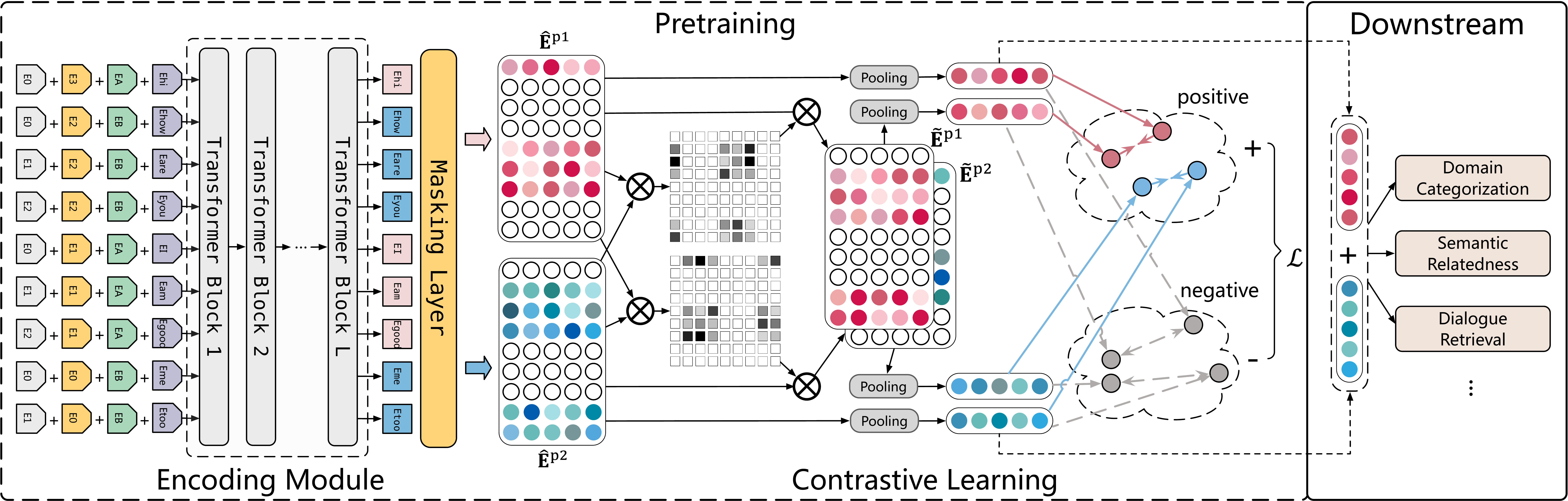}
  \caption{Architecture of dial2vec.
  Firstly, it encodes a dialogue through a PLM and assigns each interlocutor a self-representation through a masking layer (highlighted with yellow).
  Hollow circles in each self-representation represent zero embeddings.
  Then two matching matrices are calculated through the dot-product multiplication, based on which two cross-representations are generated.
  Each cross-representation and its corresponding self-representation are complementary in the token sequence dimension.
  Finally, the cosine distance between them will be minimized or maximized according to whether the training sample is positive or negative.
  }
  \label{figure:model}
\end{figure*}

\section{Related Work}
\subsection{Text Embedding}
Text embedding aims to encode a piece of text into a distributed vector that could represent its semantics.
Early works \cite{bengio2003neural, mikolov2013efficient, pennington2014glove} learn unsupervised word embeddings by making use of word-level co-occurrence information in the skip-gram or CBOW tasks.
Recently, \citet{devlin2018bert, liu2019roberta, yang2019xlnet, raffel2020exploring} pre-train deep transformer \cite{vaswani2017attention} with a series of pretext tasks, setting a new state-of-the-art across the GLUE benchmark \cite{wang2018glue} as well as exhibiting a strong potential in producing general text embeddings.
Along this line, \citet{gao2021simcse, yan2021consert, liu2021dialoguecse, chuang2022diffcse,nishikawa2022ease,zhou2022debiased,klein2022scd} fine-tune the PLMs with contrastive learning objectives, achieving remarkable improvements in learning unsupervised sentence embeddings.
\citet{luo2021unsupervised} introduce a data augmentation-based contrastive learning approach in learning document embeddings, achieving superior performance over word2vec-based approaches \cite{le2014distributed, chen2017efficient}.

For dialogue embedding, the above approaches are generally unsatisfactory, as they typically obtain dialogue embeddings by averaging the pre-trained word or sentence embeddings, ignoring the interlocutor-level conversational interactions.
Although conversational-PLMs pre-trained with dialogue data can solve this problem to some extent \cite{wu2020tod,bao2020plato,roller2021recipes}, they mainly focus on learning end-to-end models which are not sufficient for our task.
As a comparison, we study how to produce high-quality dialogue embeddings by fully exploiting the conversational information.

\subsection{Contrastive Learning}
Contrastive learning is an emerging self-supervised learning method which can improve the representation capability of PLMs in both pre-training and fine-tuning stages.
\citet{wu2020clear, meng2021coco, giorgi2020declutr} introduce the token-level and sentence-level contrastive learning tasks by correcting corrupted texts to encourage PLMs to learn noise-invariant representations. 
\citet{zhang-etal-2022-syntax} propose phrase-guided and tree-guided contrastive learning objectives to inject syntactic knowledge into PLMs.
\citet{kim2021self} propose a self-guided learning objective through which a PLM fine-tunes itself under the guidance of its different layers.
Inspired by these works, we propose to leverage the interlocutor-level conversational interactions to guide the learning of dialogue embeddings in an unsupervised learning manner.

\section{Proposed Approach}
In this section, we take a two-party dialogue as an example to describe how dial2vec works.
It is worth mentioning that dial2vec can be extended to the multi-party version through the OVR (one vs. the rest) scheme with no modification of the architecture.

\subsection{Training Samples Generation}
We first describe how we construct the positive and the negative training samples, which plays a key role in the self-guided contrastive learning approach.
Suppose that we have a dialogue dataset $\mathcal{D}=\{ S_{k} \}_{k=1}^{K}$, where $S_{k}=\{u^{p_{1}}_{1}, u^{p_{2}}_{2}, u^{p_{1}}_{3}, u^{p_{2}}_{4}, \ldots, u^{p_{1}}_{t-1}, u^{p_{2}}_{t}\}$ is the $k$-th dialogue session with $t$ utterances.
$p_{1}$ and $p_{2}$ represent two interlocutors.
We treat each utterance in a dialogue as a turn, regardless of which interlocutor it corresponds to.
For the convenience of 
narration, $k$ in $S_{k}$ is omitted in the following sections.

We treat $S$ (i.e., the original dialogue) as a positive sample.
To construct a negative sample $S'$, we first randomly select an interlocutor in $S$, say $p_{1}$, and keep all the turns of it.
Then we fill the other turns of $S$ with the utterances of $p_{2}$ randomly sampled from all dialogue sessions.
For each positive sample, we repeat this operation multiple times to generate the desired number of negative samples.

\subsection{Model Architecture}
Figure \ref{figure:model} shows the architecture of dial2vec, which consists of two parts: encoding and contrastive learning.
After training, dial2vec aggregates the embeddings from both interlocutors to obtain the final dialogue embedding, which is further used for downstream tasks. 

\subsubsection{Encoding}
Following \citet{bao2020plato}, we use four types of embeddings as input to dial2vec: token embedding, relative positional embedding, turn embedding, and role embedding.
To encode the dialogue, we first concatenate all the utterances and then tokenize them through WordPiece \cite{wu2016google} to obtain a long token sequence.
The tokens along with their corresponding position, turn, and role indices are respectively mapped into four embedding spaces and summed to form the final input embedding.

\subsubsection{Contrastive Learning}
Suppose that the output embeddings from the encoder are $\{\textbf{h}_{1}, \textbf{h}_{2}, \textbf{h}_{3}, \ldots, \textbf{h}_{n}\}$, where $\textbf{h}_{i}\in \mathbb{R}^{d}$ is the output embedding corresponding to the $i$-th input token and $n$ is the length of the input sequence,
we stack the output embeddings as a matrix denoted as $\textbf{E}\in \mathbb{R}^{n\times d}$.

To obtain the self-representations, we first generate two binary mask vectors $\textbf{m}^{p_{1}}$ and $\textbf{m}^{p_{2}}$ for two interlocutors respectively.
Let $m^{p_{1}}_{i}$ be the $i$-th element in $\textbf{m}^{p_{1}}$, then $m^{p_{1}}_{i}$ is set to 1 only when $\textbf{h}_{i}$ is derived from the input token of $p_{1}$, otherwise it is 0.
Similar operation is applied to generate $\textbf{m}^{p_{2}}$.
Then, the self-representations are obtained by:
\begin{equation}
    \label{m_vector}
    \begin{aligned}
        \hat{\textbf{E}}^{p_{1}} & = \textbf{E} \odot \left (\textbf{m}^{p_{1}}\right )^{T}, \\
        \hat{\textbf{E}}^{p_{2}} & = \textbf{E} \odot \left (\textbf{m}^{p_{2}}\right )^{T},
    \end{aligned}
\end{equation}
where $\odot$ denote the broadcast element-wise multiplication.

To extract the interaction information, we perform the token-level dot-product multiplication between $\hat{\textbf{E}}^{p_{1}}$ and $\hat{\textbf{E}}^{p_{2}}$ and compute a correlation score matrix for each interlocutor, which is formulated by:
\begin{equation}
    \label{form2}
    \begin{aligned}
        \textbf{C}^{p_{1}} & = \hat{\textbf{E}}^{p_{2}} \left (\hat{\textbf{E}}^{p_{1}}\right )^{T}, \\
        \textbf{C}^{p_{2}} & = \hat{\textbf{E}}^{p_{1}} \left (\hat{\textbf{E}}^{p_{2}}\right )^{T}, 
    \end{aligned}
\end{equation}
where $\textbf{C}^{p_{1}}$ and $\textbf{C}^{p_{2}}$ are both $n \times n$ square matrices and they are transposed to each other.
Then we generate the cross-representations by:
\begin{equation}
    \label{cross}
    \begin{aligned}
        \tilde{\textbf{E}}^{p_{1}} & = \textbf{C}^{p_{1}} \hat{\textbf{E}}^{p_{1}}, \\
        \tilde{\textbf{E}}^{p_{2}} & = \textbf{C}^{p_{2}} \hat{\textbf{E}}^{p_{2}}.
    \end{aligned}
\end{equation}

Note that $\tilde{\textbf{E}}$ can be regarded as a refined representation of $\hat{\textbf{E}}$, which highlights the conversational interaction information in the trivial encoding results.
The fact that $\tilde{\textbf{E}}$ and $\hat{\textbf{E}}$ share the same semantic space allows us to directly optimize their cosine distance without any additional transformations.
In this circumstance, $\tilde{\textbf{E}}$ acts as guidance for leading $\hat{\textbf{E}}$ to be an interaction-aware self-representation, and this is why we call dial2vec works in a self-guided manner.

We further introduce $w$ as a restriction hyper-parameter to mask the long-range semantic correlations among the utterances of $p_{1}$ and $p_{2}$.
Specifically, let $\textbf{C}[i, j]$ denotes the element in the $i$-th row and the $j$-th column of $\textbf{C}$ in Eq. (\ref{form2}).
$T(i)$ represents a function which returns the turn index for the $i$-th output embedding in $\textbf{E}$.
Then $\forall{i, j} \in {1, 2, \ldots, n}$, $\textbf{C}[i, j]$ is masked with zero where $abs(T(i) - T(j)) > w$, otherwise remains unchanged.
Here we omit the superscript $p_{1}$ and $p_{2}$ in $\textbf{C}$ since they are processed in the same way.

\subsubsection{Aggregation}
\label{sec:dialogue embedding}
To obtain the dialogue embedding $\textbf{e}$, we compare two aggregation strategies.
In the first strategy, we directly perform average pooling across all entire output embeddings $\textbf{E}$ (here we do not distinguish between $p_{1}$ and $p_{2}$).
We further propose the interlocutor-level pooling strategy, formulated as:

\begin{equation}
\label{eq:aggregation strategy}
    \begin{aligned}
    \textbf{e} & = \sum_{r=1}^{R}\dfrac{\sum_{i=1}^{n} m^{p_{r}}_{i} \textbf{h}_{i}}{\sum_{i=1}^{n} m^{p_{r}}_{i}},
    \end{aligned}
\end{equation}
where $m_{i}^{p_{r}}$ is the $i$-th value in $\textbf{m}^{p_{r}}$ and $R$ is the number of interlocutors.
We compare the results of the two strategies in Section \ref{section:aggregation strategy}.

\subsubsection{Learning Objective}
We adopt the NT-Xent loss proposed in \cite{oord2018representation} to train our model.
Let $N$ be the number of all training samples associated with $S$, which actually equals one positive sample plus the number of its corresponding negative samples. The loss $l$ is defined as:

\begin{equation}
    \begin{aligned}
       l & = -\sum_{r=1}^{R}\log\frac{ e^{\text{sim}(\hat{\textbf{E}}^{p_{r}}, \tilde{\textbf{E}}^{p_{r}})/\tau} }{\sum_{j=1}^{N}e^{\text{sim}(\hat{\textbf{E}}^{p_{r}}_{j}, \tilde{\textbf{E}}^{p_{r}}_{j})/\tau }}, 
    \end{aligned}
\end{equation}
where $\tau$ is the hyper-parameter of temperature.
$\text{sim}(\cdot,\cdot)$ is defined as an average pooling operation followed by the cosine distance calculation.
For all $K$ dialogues in the dataset $D$, the loss $\mathcal{L}$ is given by:
\begin{equation}
    \begin{aligned}
        \mathcal{L} &= \frac{1}{K}\sum_{i=1}^{K} l_{i}. \\
    \end{aligned}
\end{equation}

\label{sec: datasets}
\begin{table*}[tb]
\tabcolsep=5.5pt
\setlength{\belowcaptionskip}{-.4cm}
    \centering
    \footnotesize
    \renewcommand{\arraystretch}{1.1}
    \begin{tabular}{c|ccc|ccc|ccc|c}
	\toprule
        \multirow{2}{*}{Datasets} & \multicolumn{3}{c|}{Train} & \multicolumn{3}{c|}{Dev} & \multicolumn{3}{c|}{Test} & \multirow{2}{*}{\#Domain} \\
        \cline{2-10}
		 & \#Sample & \#Turn & \#Word & \#Sample & \#Turn & \#Word & \#Sample & \#Turn & \#Word &  \\
	\midrule
        BiTOD           & 2952   & 19 & 217    & 70    & 11 & 109  & 106   & 10     & 106 & 6 \\
		Doc2dial        & 3474  & 11 & 187    & 661   & 12 & 182  & 661   & 12    & 182 & 4 \\
		MetalWOZ        & 30307 & 11 & 83    & 3788  & 11 & 82   & 3789  & 11     & 82  & 47 \\
		MultiWOZ        & 8437 & 13 & 177   & 1077   &  9 & 110  & 1084   &  9     & 110 & 7\\
		Self-dialogue   & 19331 & 15 & 151    & 2416  & 15 & 151  & 2417  & 15     & 152 & 28 \\
		SGD             & 16142 & 20 & 199    & 836   & 14 & 140  & 1331  & 12     & 124 & 45 \\

	\bottomrule
    \end{tabular}
\caption{
Statistics of the six dialogue datasets used in our experiments.
\#Turn and \#Word represent the average number of turns and words per dialogue.
\#Domain represents the total number of domains in the dataset.
}
\label{tab:dataset_profile}
\end{table*}

\section{Experiments Setup}
\label{section4}
\subsection{Evaluation Tasks}
We introduce three evaluation tasks: domain categorization, semantic relatedness, and dialogue retrieval.
We categorize them into intrinsic and extrinsic tasks.
The intrinsic tasks, including domain categorization and semantic relatedness, focus on assessing the overall distribution of the learned dialogue embeddings.
The extrinsic task (i.e., dialogue retrieval) is more concerned with the performance of embeddings in dense retrieval scenarios.

\textbf{Domain Categorization}.
Given a dataset of dialogues, the task is to assign each dialogue the corresponding domain label.
Following \citet{schnabel2015evaluation}, we conduct this experiment as an unsupervised clustering task.
All the dialogue embeddings are clustered into $n$ categories with KMeans++ Algorithm \cite{arthur2006k}, where $n$ is the number of domains in the dataset.
We adopt the purity metric in this task.

\textbf{Semantic Relatedness}.
We pair each dialogue with a dialogue randomly selected from the same dataset and evaluate their semantic relatedness score based on their cosine similarity.
The ground-truth label assigned to each dialogue pair is a binary value and is decided by whether the two dialogues share the identical domain.
Following \citet{baroni2014don}, we calculate Spearman's correlation between the sorted semantic relatedness scores and their corresponding ground-truth labels.
This task is more stable than the domain categorization task since it gets rid of the high variance of clustering algorithms when the embedding distribution changes.

\textbf{Dialogue Retrieval}.
Given a dialogue as a query, this task requires a model to rank all the candidates based on the cosine similarities.
We use mean average precision (MAP) as the evaluation measure.

\subsection{Datasets}
We collect six widely-used dialogue datasets as below.
We choose these datasets because they hold clear domain labels.
Other datasets either provide non-semantic labels (e.g., logical labels that are less relevant to conversational semantics) \cite{li2017dailydialog} or provide the domain labels automatically annotated by algorithms \cite{chen2021dialogsum}, thus are not suitable in our experiments.
We split each dataset into training, validation, and testing sets and filter out dialogues with multiple domains in validation and test sets to fit our evaluation tasks.
All domain labels are invisible to the model during the training procedure.
Table \ref{tab:dataset_profile} shows their statistics.

\textbf{BiTOD} \cite{lin2021bitod} is a bilingual multi-domain dataset proposed for end-to-end task-oriented dialogue modeling.
It provides thousands of dialogues and a large and realistic bilingual knowledge base.
We use the dialogues and conduct experiments under the monolingual setting.

\textbf{Doc2dial} \cite{feng2020doc2dial} includes goal-oriented dialogues that are grounded in the associated documents.
We take the document topics as the domain labels of the dialogues.

\textbf{MetalWOZ} \cite{lee2019multi-domain} is proposed for DSTC8, aiming at helping models more accurately predict user responses in new domains.

\textbf{MultiWOZ} \cite{eric2019multiwoz} is a multi-domain dialogue dataset that poses significant challenges to task-oriented dialogue modeling due to its complexity.
We use the 2.1 version in our experiments.

\textbf{Self-dialogue} \cite{fainberg2018talking} consists of large-scale self-dialogues with a broad set of topics.
Modeling these dialogues is relatively difficult since they have more turns and topics.

\textbf{SGD} \cite{rastogi2020towards} is another larger-scale multi-domain dialogue dataset.
We take the service field as the domain label of the dialogues.

\begin{table*}[thb]
\tabcolsep=5.5pt
\setlength{\belowcaptionskip}{-.4cm}
    \centering
    \small
    \begin{tabular}{lcccccccccccccc}
        \toprule
        \multirow{2}{*}{\textbf{Model}} & \multicolumn{7}{c}{\textbf{Domain Categorization}} & \multicolumn{7}{c}{\textbf{Semantic Relatedness}} \\
        \cmidrule(lr){2-8}\cmidrule(lr){9-15}
        & \textbf{bit} & \textbf{doc} & \textbf{met} & \textbf{mul} & \textbf{sel} & \textbf{sgd} & \textbf{Average} & \textbf{bit} & \textbf{doc} & \textbf{met} & \textbf{mul} & \textbf{sel} & \textbf{sgd} & \textbf{Average} \\
        \midrule
         LDA       & 44.7             & 35.2 & 19.3 & 45.9 & 24.7 & 20.2 & 31.7 & - & - & - & - & - & - & - \\
        \midrule
         GloVe & 64.3             & 54.7 & 40.5 & 79.0 & 35.0 & 51.6 & 54.2 & 34.3 & 25.9 & 15.8 & 38.9 & 15.7 & 27.6 & 26.4 \\
         Doc2Vec  & 82.7 & 70.9 & 43.9 & 86.1 & 40.9 & 63.6 & 64.7 & 43.4 & 24.8 & 14.6 & 30.6 & 10.7 & 26.9 & 25.2 \\
         SimCSE   & 79.3  & 64.7 & 45.1 & 85.5 & 46.8 & 66.7 & 64.7 & 39.5 & 33.2 & 14.9 & 38.1 & 18.3 & 26.9 & 28.5 \\
         DialogueCSE & \underline{85.8}  & 68.4 & \underline{77.5} & \underline{94.9} & 53.2 & 72.1 & \underline{75.3} & 42.4 & \underline{44.5} & 23.9 & \underline{65.2} & 27.6 & 31.7 & \underline{39.2} \\

        \midrule
        BERT    & 49.1 & 54.0 & 31.6 & 61.3 & 44.4 & 31.3 & 45.3 & 24.3 & 22.4 & 11.6 & 30.5 & 16.7 & 17.9 & 20.6 \\
        RoBERTa & 63.2 & 40.4 & 46.4 & 62.8 & 44.9 & 40.8 & 49.8 & 30.2 & 14.8 & 15.4 & 28.5 & 17.5 & 16.7 & 20.5 \\
        T5      & 78.7 & 55.2 & 67.6 & 89.5 & 43.8 & 69.5 & 67.4 & 38.6 & 28.6 & 20.8 & 42.5 & 20.0 & 29.7 & 30.0 \\

        \midrule
        TOD-BERT  & 75.6 & 63.1  & \textbf{82.9}  & 94.3 & 50.0 & 50.3 & 69.4 & \underline{47.0} & 32.6 & \underline{24.3} & 48.9 & 24.6 & 24.8 & 33.7 \\
        Blender   & 80.9 & 56.4             & 62.3             & 82.4             & 45.4 & \underline{73.1} & 66.8 &  37.0 & 28.1 & 19.9 & 44.4 & 18.3 & 31.1 & 29.8 \\
        PLATO     & 74.1 & \underline{79.0} & 73.9             & 82.5             & \underline{62.5} & 71.0 & 73.8 & 46.6 & 38.7 & 22.7 & 45.3 & \underline{35.1} & \underline{32.4} & 36.8 \\
        \midrule
        Dial2Vec & \textbf{90.6} & \textbf{90.2} & 77.2 & \textbf{96.7} & \textbf{63.1} & \textbf{86.2} & \textbf{84.0} & \textbf{68.8} & \textbf{50.7} & \textbf{24.5} & \textbf{71.0} & \textbf{37.2} & \textbf{36.9} & \textbf{48.2}\\
        
        \bottomrule
    \end{tabular}
    \caption{
    Evaluation results of the intrinsic tasks on the six dialogue datasets, including BiTOD (\textbf{bit}), Doc2dial (\textbf{doc}), MetalWOZ (\textbf{met}), MultiWOZ (\textbf{mul}), Self-dialogue (\textbf{sel}) and SGD (\textbf{sgd}).
    The metrics are purity and Spearman's correlation for the two tasks respectively.
    All results reported are averaged across 10 independent runs to reduce the variance.
    Boldface and underline highlight the best and the second-best scores.
    }
    \label{tab:intrinsic}
\end{table*}

\subsection{Comparison Methods}
\label{sec: comparison methods}
The baseline approaches compared to our model are categorized into four groups as follows.

\textbf{Non-DL Approaches}.
We treat a dialogue as a document and apply Latent Dirichlet Allocation (LDA) \cite{blei2003latent} to assign each dialogue a topic.
LDA is only used in the domain categorization task, since it cannot give a similarity score between two dialogues.

\textbf{Embedding-based Approaches}.
We consider a dialogue as a sequence of words or sentences, and we obtain dialogue embeddings by combining their pre-trained embeddings.
We adopt GloVe \cite{pennington2014glove} to obtain pre-trained word embeddings, SimCSE \cite{gao2021simcse} to obtain pre-trained universal sentence embeddings, and DialogueCSE \cite{liu2021dialoguecse} to obtain pre-trained dialogue-based sentence embeddings.
Also, we consider a dialogue as a document and embed it with Doc2Vec \cite{le2014distributed}.

\textbf{PLMs}.
We consider three representative pre-trained language models, including BERT \cite{devlin2018bert}, RoBERTa \cite{liu2019roberta}, and T5 \cite{raffel2020exploring}.

\textbf{Conversational-PLMs}.
We adopt TOD-BERT \cite{wu2020tod}, Blender \cite{roller2021recipes}, and PLATO \cite{bao2020plato} as baselines in this group.
TOD-BERT \cite{wu2020tod} is pre-trained with nine dialogue datasets, including MultiWOZ and MetalWOZ.
We adopt it as a strong baseline to compare against our model, especially on the MultiWOZ and MetalWOZ datasets.
Blender \cite{roller2021recipes} and PLATO \cite{bao2020plato} are pre-trained with large-scale open domain dialogue data including Twitter and Reddit \cite{cho2014learning, zhou2018commonsense,galley2019grounded}.
We also include them as strong baselines for comparison.

For all PLMs and Conversational-PLMs, we use the average of the output embeddings from the top layer as the dialogue embedding.
We do not experiment with the \texttt{[CLS]} token embedding since it may be relatively weak in representing long conversational texts.

\subsection{Implement Details}
Our model is implemented in PyTorch \cite{paszke2019pytorch}.
We initialize our encoder with PLATO's pre-trained parameters.
During fine-tuning, we freeze the bottom 6 layers of the encoder to avoid the catastrophic forgetting problem.
The maximum sequence length is limited to 512 but is sufficient for most dialogues in our experiments.
The temperature $\tau$ and the window size $w$ 
are set to 0.2 and 10 respectively, since such configuration performs best across all datasets.
We optimize the model parameters with Adam optimizer \cite{kingma2014adam}, using a learning rate of 1e-5 and a batch size of 5 per GPU.
All models are trained with 4 NVIDIA Tesla V100 GPUs.

\section{Experimental Results}
\subsection{Intrinsic Task}
Table \ref{tab:intrinsic} shows the experimental results of the intrinsic tasks.
For each task, dial2vec achieves on average 8.7 and 9.0 absolute improvements in terms of purity and Spearman's correlation against the strongest baseline DialogueCSE.
We attribute the strong performance to the introduction of the interlocutor-level conversational interaction information in learning dialogue embeddings.

Conversational-PLMs show overwhelming superiority over PLMs, indicating that pre-training with conversational data plays a key role in producing better dialogue embeddings.
The phenomenon that TOD-BERT achieves very competitive results against dial2vec on MultiWOZ and MetalWOZ also confirms this fact.
Even so, dial2vec easily bridges or reverses the gaps between PLATO and TOD-BERT on both datasets, demonstrating its superiority in exploiting conversational information.

PLATO generally performs better than TOD-BERT and Blender.
We hypothesize that the turn and role embeddings also play a crucial role in our task.
To verify this, we employ BERT as the encoder and train dial2vec(BERT)\footnote{In this case, \texttt{[EOU]} tokens are inserted into the sequence to denote the separation of different turns.} under the same setting as dial2vec(PLATO).
However, the results on the three tasks degrade rapidly after reaching the best performances.
We believe that under such a setting, the inputs provide insufficient information for the model to maintain turn-aware and role-aware semantic information in the encoding outputs, making the training not robust.

The embedding-based methods suffer from poor performance since they ignore the weights when combining word and sentence embeddings.
Among them, SimCSE releases the inherent representation capability of PLMs by introducing the twice-dropout operation in fine-tuning, achieving better results than GloVe and Doc2Vec.
But since such an operation is generic and orthogonal to our work, we do not incorporate it into our model.
Particularly, DialogueCSE yields superior results compared with other embedding-based methods and even shows competitive performances against our model.
This is reasonable since it is the only baseline in this group that leverages conversational interactions to learn sentence embeddings.
However, its performance is still unsatisfactory since it performs sentence-level instead of interlocutor-level interactions and fails to model the entire dialogue.

\subsection{Extrinsic Task}
Table \ref{tab:extrinsic} shows dial2vec's performances on the dialogue retrieval task.
Compared to the experiment results on the intrinsic tasks, dial2vec achieves more significant improvements on all datasets.
We attribute it to dial2vec's capability of understanding fine-grained conversational semantics.
Since dial2vec is forced to distinguish the positive samples composed of the exact matching question-answers from the negative ones, the semantic information it learned is more fine-grained than tasks trained with only domain labels.
Such characteristic makes it adept at ranking semantically similar candidates, resulting in better performance on the MAP metric.

\begin{table}[bht]
\tabcolsep=4pt
\setlength{\belowcaptionskip}{-.4cm}
    \centering
    \small
    \begin{tabular}{lccccccc}
        \toprule
        \multirow{2}{*}{\textbf{Model}} & \multicolumn{7}{c}{\textbf{Dialogue Retrieval}}  \\
        \cmidrule(lr){2-8}
        & \textbf{bit} & \textbf{doc} & \textbf{met} & \textbf{mul} & \textbf{sel} & \textbf{sgd} & \textbf{AVG} \\
        \midrule
         GloVe & 63.8 & 49.1 & 29.4 & 65.9 & 24.1 & 52.6 & 47.5 \\
         Doc2Vec   & 67.7 & 43.7 & 15.1 & 50.9 & 16.3 & 43.1 & 39.5 \\
         SimCSE    & 62.5 & 52.5 & 23.8 & 62.1 & 27.0 & 44.8 & 45.5 \\
         DialogueCSE  & 72.9 & 58.2 & \underline{66.7} & 82.9 & 34.5 & 62.5 & \underline{62.9} \\

        \midrule
         BERT    & 52.4 & 44.8 & 17.0 & 56.4 & 25.8 & 26.0 & 37.1  \\
         RoBERTa   & 62.2 & 40.6 & 30.4 & 57.4 & 25.5 & 35.0 & 41.9 \\
         T5        & 67.3 & 49.9 & 43.9 & 69.7 & 27.8 & 53.8 & 52.1 \\

        \midrule
        TOD-BERT & \underline{73.2} & 53.0 & 65.7 & \underline{84.2} & 33.0 & 45.3 & 59.1 \\
        Blender & 69.1 & 50.1 & 44.6 & 70.1 & 25.4 & 63.0 & 53.7 \\
        PLATO & 71.6 & \underline{59.7} & 54.5 & 68.7 & \underline{45.9} & \underline{63.2} & 60.6 \\
        
        \midrule
        Dial2Vec & \textbf{94.4} & \textbf{69.4} & \textbf{68.0} & \textbf{96.4} & \textbf{49.4} & \textbf{82.8} & \textbf{76.7} \\

        \bottomrule
    \end{tabular}
    \caption{
    Evaluation results of the dialogue retrieval task.
    We use the mean average precision (MAP) as the evaluation metric.
    Boldface and underline highlight the best and the second-best scores.
    }
    \label{tab:extrinsic}
\end{table}

\subsection{Analysis}
To further investigate the property of our model, we adopt PLATO as the baseline to conduct experiments with the single interlocutor's embeddings, the aggregation strategies, and the embedding distributions.
We report the average results across all datasets.

\subsubsection{Single Interlocutor's Embeddings}
Intuitively, each interlocutor holds their own unilateral information of the dialogue.
However, such interaction-free information usually contains noises or overlaps with that from other interlocutors.
Table \ref{tab:ablation_role} shows the experiment results of PLATO and dial2vec.
We find that the PLATO's embeddings for individual interlocutors usually perform close to or even better than the aggregated results.
As a comparison, dial2vec yields significantly better embeddings for both interlocutors, and achieves further improvements when aggregating them.
We conclude that under the guidance of the conversational interactions, dial2vec eliminates the interlocutor-level interaction-free information and highlights the interaction-aware information, thus achieving better performances.


        
         
        
         
         
         

\begin{table}[htb]
\setlength{\belowcaptionskip}{-.4cm}
\tabcolsep=4pt
    \centering
    \small
    \renewcommand{\arraystretch}{1}
    \begin{tabular}{l|c|ccc}
        \toprule 
         \textbf{Model}  &  \textbf{Interlocutor}  & \textbf{Purity} & \textbf{Spearman}  & \textbf{MAP}  \\

        \midrule
        \multirow{3}{*}{PLATO} & $p1$ & 71.78 & 36.23 & 59.38 \\
                               & $p2$ & 75.29 & 36.67 & 60.73 \\
                               & $p1+p1$ & 73.83 & 36.80 & 60.60 \\
        \cmidrule{2-5}
        & diff & -1.46 & 0.13 & -0.13 \\
        
        \midrule
        \multirow{3}{*}{Dial2Vec} & $p1$ & 83.27 & 47.82 & 75.72 \\
                                  & $p2$ & 82.21 & 46.74 & 74.56 \\
                                  & $p1+p1$ & \textbf{83.97} & \textbf{48.19} & \textbf{76.73} \\
        \cmidrule{2-5}
        & diff & 0.70 & 0.37 & 1.01 \\
                                  
        \bottomrule
    \end{tabular}
    \caption{
    Performances of dialogue embeddings for each interlocutor.
    $p1$ represents that we use the embeddings from the interlocutor who starts the conversation, and $p2$ represents the opposite.
    $p1+p2$ represents the aggregated results.
    $diff$ shows improvements of the aggregated results over the best single interlocutor's results.
    Boldface represents the best scores among the models.}
    \label{tab:ablation_role}
\end{table}

\subsubsection{Aggregation Strategies}
\label{section:aggregation strategy}
As described in Section \ref{sec:dialogue embedding}, we experiment with two aggregation strategies: average pooling and interlocutor-level pooling.
For average pooling, we average all the output embeddings as the final dialogue embedding, while for interlocutor-level pooling, we sum the average pooling results of the output embeddings corresponding to each interlocutor.

\begin{table}[htb]
    \centering
    \small
    \renewcommand{\arraystretch}{1}
    \begin{tabular}{l|ccc}
        \toprule 
        \textbf{Model}  & \textbf{Purity} & \textbf{Spearman}  & \textbf{MAP}  \\

        \midrule

         Dial2Vec$_{avg}$ & 83.69 & 47.93 & 76.39 \\
         Dial2Vec$_{int}$ & \textbf{83.97} & \textbf{48.19} & \textbf{76.73} \\
         diff & +0.28 & +0.26 & +0.34 \\  

        \bottomrule
    \end{tabular}
    \caption{
    Comparison between the average-pooling (denoted as ${avg}$) and interlocutor-level pooling (denoted as ${int}$) strategies.
    Boldface highlights the best scores.
    }
    \label{tab:aggregation strategy}
\end{table}

Table \ref{tab:aggregation strategy} shows the results for the two strategies on all datasets.
The interlocutor-level pooling strategy performs consistently better than the average pooling strategy.
We hold that the interlocutor-level pooling strategy acts as a normalization operation that balances the weight of semantic information from different interlocutors.

\subsubsection{Alignment and Uniformity Analysis}
\label{section:alignment and uniformity}
Inspired by \citet{wang2020understanding}, we employ the alignment and uniformity metrics to study the variation of dialogue embedding distribution during training.
Given a set of data pairs and their corresponding labels, the alignment metric is calculated as the expected value of Euclidean distances of each positive data pair, formulated as:
\begin{equation}
\label{definition of alignment}
    \ell_{alignment} \triangleq \mathop{\mathbb{E}}\limits_{x,x^{+}\sim p_{pos}} \Vert f(x) - f(x^{+})\Vert^{2}.
\end{equation}

\vspace{-0.25cm}

The alignment metric is suitable for tasks such as \citet{gao2021simcse} since the positive pairs are encoded from a unique text.
However, in our scenario, positive pairs generated from two different dialogues are only expected to have closer distances than negative pairs.
Thus, we revise the Eq. (\ref{definition of alignment}) to be:
\begin{equation}
    \begin{aligned}
        \ell_{adj\_alignment} & \triangleq \mathop{\mathbb{E}}\limits_{x,x^{+}\sim p_{pos}} \Vert f(x) - f(x^{+})\Vert^{2}, \\
        & - \mathop{\mathbb{E}}\limits_{x,x^{-}\sim p_{neg}} \Vert f(x) - f(x^{-})\Vert^{2}.
    \end{aligned}
\end{equation}
We name $\ell_{adj\_alignment}$ as the adjusted alignment metric.

The uniformity metric is defined to measure how close the embeddings are to the uniform distribution:
\begin{equation}
    \ell_{uniformity} \triangleq \log \mathop{\mathbb{E}}\limits_{x,y\mathop{\sim}\limits^{i.i.d} p_{data}} e^{-2\Vert f(x) - f(y)\Vert^{2}},
\end{equation}
where $p_{data}$ denotes the data distribution.

\begin{figure}[thb]
\setlength{\belowcaptionskip}{-.4cm}
  \centering
  \includegraphics[width=1\linewidth]{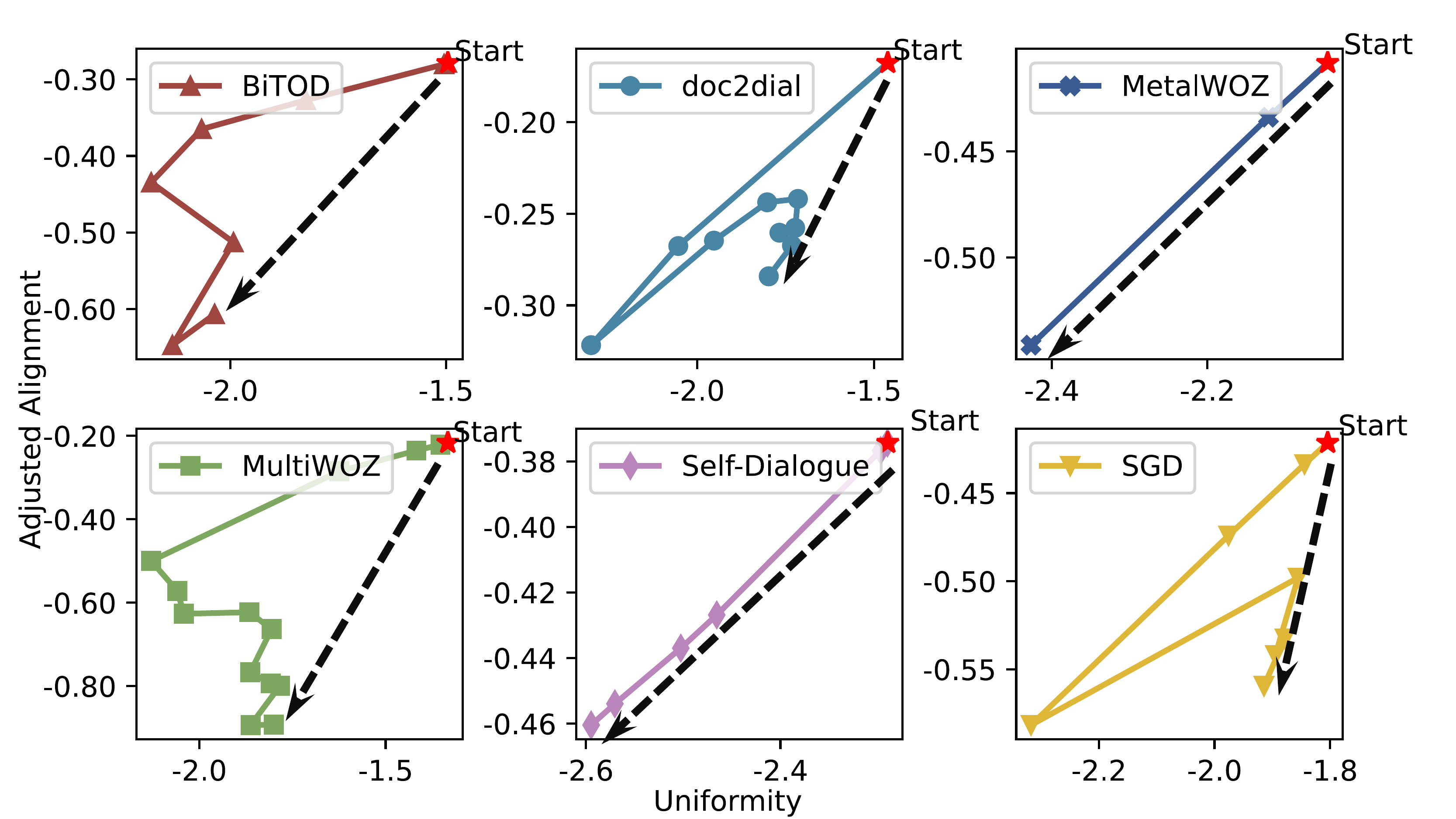}
  \caption{
  The scatter plot of $\ell_{adj\_alignment}$-$\ell_{uniformity}$ on the six testing sets.
  We plot the two metrics after every evaluation.
  For both metrics, lower values represent better distributions.
  }
  \label{fig:align_uniform}
\end{figure}

Figure \ref{fig:align_uniform} shows how the adjusted alignment and uniformity vary with the training iterations.
`Start' marks the results at the very beginning of training, which also stands for the vanilla PLATO's performances.
As we can see, the two metrics decrease rapidly in the first few iterations.
We believe dial2vec learns discriminative embeddings by pushing the embeddings for the two interlocutors in the negative samples away from each other in this stage.
Since the dialogue embeddings are spread out over the unit hypersphere, both metrics decrease.
As the training proceeds, the model learns the fine-grained informative dialogue embeddings from the positive samples.
This makes the dialogues with similar semantics close to each other, causing the uniformity to increase again.
The two metrics finally converge to the values much better than the start points, showing that dial2vec learns both informative and discriminative embeddings.

\section{Conclusion}
In this paper, we formally introduce the task of learning unsupervised dialogue embeddings and propose dial2vec to solve this task.
We introduce a self-guided mechanism that leverages the conversational interactions to guide the learning of the embeddings for both interlocutors and propose the interlocutor-level strategy to aggregate them.
We further release a benchmark consisting of six widely-used dialogue datasets and three tasks designed based on the domain labels.
Our model achieves superior performances on all datasets across the three tasks, and further analysis shows that the dialogue embeddings learned by our model are more informative and discriminative than the baselines.
We believe there is still much room for improvement to generate satisfactory dialogue embedding.

\section{Limitations}
Our work has two limitations.
First, although dial2vec is designed to be able to scale to multi-party conversations, we did not conduct such experiments due to the lack of a suitable multi-party evaluation dataset.
As annotating for a multi-party dialogue dataset is indeed complicated, we leave it to future work.
Besides, dial2vec still performs unsatisfactorily when employing BERT-like encoders.
Although they also achieve very competitive results, we believe that a robust training procedure is more important since we do not know when to stop training under the unsupervised setting in practice.
Dial2vec should be further improved to better adapt to the multiple formats of input embeddings.


\bibliography{acl_latex}
\bibliographystyle{acl_natbib}



\end{document}